\documentclass[10pt,twocolumn,letterpaper]{article}

\usepackage{iccv}              

\usepackage{multirow}
\usepackage{arydshln}
\usepackage{stmaryrd}
\usepackage{lipsum}
\usepackage{graphicx}
\usepackage{caption}

\definecolor{iccvblue}{rgb}{0.21,0.49,0.74}
\usepackage[pagebackref,breaklinks,colorlinks,allcolors=iccvblue]{hyperref}

\def\paperID{5760} 
\def\confName{ICCV}
\def\confYear{2025}

\title{RARE: Refine Any Registration of Pairwise Point Clouds via Zero-Shot Learning}

\author{Chengyu Zheng$^{1,2}\footnotemark[1]$ \quad  Jin Huang$^{1}\footnotemark[1]$  \quad Honghua Chen$^3\footnotemark[2]$ \quad Mingqiang Wei$^{1,2}\footnotemark[2]$ \\
$^1$College of Computer Science and Technology, Nanjing University of Aeronautics and Astronautics \\
$^2$Shenzhen Research Institute, Nanjing University of Aeronautics and Astronautics\\ 
$^3$School of Data Science, Lingnan University, Hong Kong SAR, China\\ 
{\tt\small zhengcy@nuaa.edu.cn; jinhuang.nuaa@gmail.com; chenhonghuacn@gmail.com; mqwei@nuaa.edu.cn}
}

\begin{document}

\twocolumn[{
\renewcommand\twocolumn[1][]{#1}
\maketitle
\begin{center}
    \captionsetup{type=figure}
    \includegraphics[width=\textwidth]{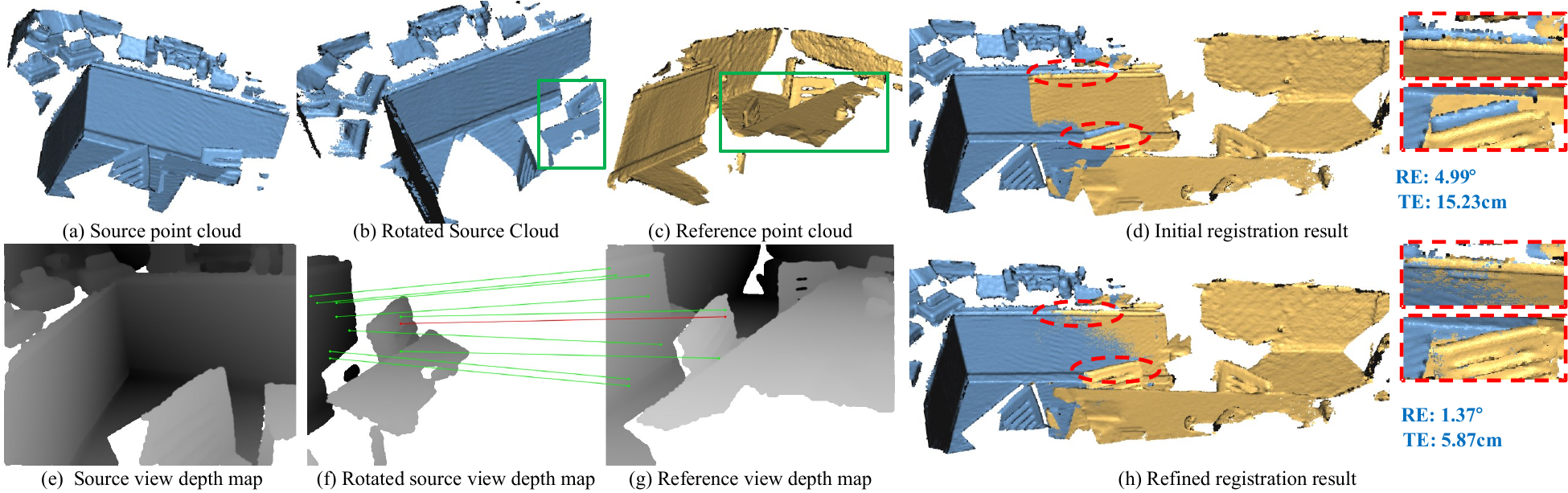}
    \captionof{figure}{
    RARE refines arbitrary point cloud registration methods effectively in a zero-shot learning manner.
    (a) and (e) show the source point cloud and its projected depth map from the source view.
    (b) and (f) depict the source point cloud rotated to the reference view using the initial transformation $T_{init}$ and its corresponding projected depth map.
    (c) and (g) display the reference point cloud and its projected depth map from the reference view. 
    From (f) and (g), it is evident that the initial transformation allows us to extract high-quality correspondences from the depth maps.
    (d) shows the registration result using the initial transformation by GeoTransformer \cite{qin2023geotransformer}, while (f) presents the refined registration result by diffusion features.
    }
    \label{fig:teaser}
\end{center}
}]

\footnotetext[1]{Equal Contribution} 
\footnotetext[2]{Corresponding Author}

\begin{abstract}

Recent research leveraging large-scale pretrained diffusion models has demonstrated the potential of using diffusion features to establish semantic correspondences in images.
Inspired by advancements in diffusion-based techniques, we propose a novel zero-shot method for refining point cloud registration algorithms.
Our approach leverages correspondences derived from depth images to enhance point feature representations, eliminating the need for a dedicated training dataset.
Specifically, we first project the point cloud into depth maps from multiple perspectives and extract implicit knowledge from a pretrained diffusion network as depth diffusion features.
These features are then integrated with geometric features obtained from existing methods to establish more accurate correspondences between point clouds.
By leveraging these refined correspondences, our approach achieves significantly improved registration accuracy.
Extensive experiments demonstrate that our method not only enhances the performance of existing point cloud registration techniques but also exhibits robust generalization capabilities across diverse datasets.
Codes are available at \href{https://github.com/zhengcy-lambo/RARE.git}{https://github.com/zhengcy-lambo/RARE.git}.
\end{abstract}    
\section{Introduction}
Point cloud registration is a fundamental yet challenging task in the field of 3D computer vision, with broad applications in localization~\cite{fischler1981random}, object detection~\cite{guo20143d}, and scene reconstruction~\cite{mian2005automatic}.
The objective is to compute a rigid transformation that aligns two partial point clouds captured from different viewpoints of the same scene or object.

Point cloud registration typically comprises three key steps: feature extraction, correspondence establishment, and robust geometric transformation estimation.
Feature extraction is a critical component, as distinctive and unique features enable the establishment of more accurate correspondences, thereby improving registration accuracy.
Traditional methods~\cite{rusu2009fast, johnson1999using} often rely on hand-crafted features, which tend to exhibit limited performance.
Recently, rapid advancements in 3D representations have driven the development of learning-based 3D features for point cloud registration.
Learning-based methods~\cite{choy2019fully,huang2021predator,yu2021cofinet,yew2022regtr,wang2022you,qin2023geotransformer} extract 3D features using feature extractors trained on large datasets of paired point clouds.
While these approaches demonstrate robust performance under challenging conditions, such as low overlap, they require substantial amounts of paired data with correspondences for training, limiting their practical applicability.
However, recent advancements in pre-trained generative models offer promising alternatives that reduce the dependency on paired data.
For instance, Stable Diffusion (SD)~\cite{rombach2022high} exhibits remarkable capabilities in synthesizing high-quality images from textual input, highlighting its powerful internal representation of images.
Diffusion-based methods~\cite{tang2023emergent, zhang2024tale} leverage SD features to establish semantic correspondences across images.
Building on this, FreeReg~\cite{wang2023freereg} extends the approach by establishing robust correspondences between images and point clouds using pre-trained diffusion models.

\begin{figure}[t]
  \centering
  \includegraphics[width=\linewidth]{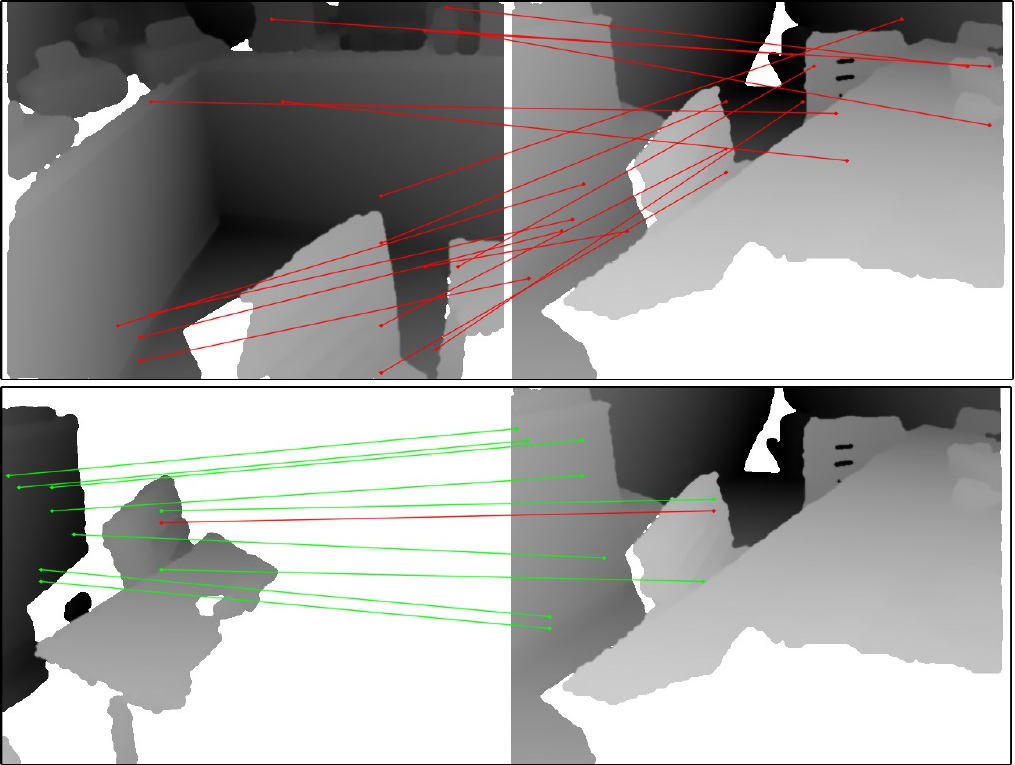}
  \caption{Comparison of correspondence matching using projected depth maps from different viewing angles.
  In the top pair of images, the default viewing angle introduces challenges in obtaining accurate correspondences from depth maps due to the reliance on diffusion features.
  In contrast, in the bottom pair of images, aligning both point clouds to a consistent viewing angle enables the projected depth maps to produce more precise correspondences.
  Incorrect correspondences are indicated by red lines, while correct correspondences are marked by green lines.
  }
  \label{fig:limitation}
\end{figure}

Inspired by diffusion-based approaches, we explore the use of diffusion model features to establish correspondences between two point clouds.
A straightforward method for converting point clouds into an image modality involves projecting the points onto an image plane to create a depth map, which can then be transformed into an image using depth-to-image diffusion models, such as ControlNet~\cite{zhang2023adding}.
However, as highlighted by FreeReg, a single depth map can correspond to multiple plausible images, leading to significant discrepancies between the generated images and the original input.
To address this, FreeReg leverages intermediate feature maps from the depth-to-image ControlNet to directly establish correspondences between depth maps derived from point clouds.
Nevertheless, existing methods like FreeReg face significant challenges in generating accurate correspondences on depth maps with large parallax, as illustrated in Figure \ref{fig:limitation}.
To mitigate this issue, we propose using an initial transformation derived from existing point cloud registration methods to convert matched depth maps from large parallax to small parallax.
This approach is based on our observation that diffusion features on depth maps with smaller parallax exhibit higher distinctiveness.
By leveraging these diffusion features from small parallax depth maps, we refine existing methods to achieve more accurate rigid transformations.

In this paper, we propose a zero-shot approach to refine existing point cloud registration methods by leveraging correspondences derived from depth images, enhancing point feature representations without requiring a training dataset.
First, we project the point cloud into depth maps using the initial transformation obtained from existing point cloud registration methods and extract implicit knowledge as depth diffusion features using pre-trained diffusion models.
These features are then integrated with the 3D features extracted from existing methods to establish more accurate correspondences between point clouds.
To evaluate our approach, we conduct experiments on three datasets: the scene-scale indoor datasets 3DMatch and 3DLoMatch~\cite{zeng20173dmatch}, and the scene-scale outdoor dataset KITTI~\cite{andreas2012are}.
Our method not only improves the performance of existing point cloud registration techniques but also demonstrates robust generalization capabilities across diverse datasets.
To summarize, our contributions are as follows:
\begin{itemize}
\item 
We propose a zero-shot approach to refine point cloud registration using depth information, eliminating the need for training on point cloud datasets.  
\item 
We leverage pre-trained diffusion models to extract deep features from depth maps and integrate them with 3D features derived from point clouds through a series of aggregation operations, enabling the establishment of more accurate correspondences.
\item We conduct extensive experiments on three benchmarks such as 3DMatch, 3DLoMatch, and KITTI, and achieve competitive results.
\end{itemize}


\section{Related Work}
\subsection{Point Cloud Registration}

Point cloud registration aims to estimate a rigid transformation that aligns two partially overlapping 3D point clouds and is widely utilized in graphics, vision, and robotics.
The traditional approach focuses on hand-crafted descriptors that can discriminatively characterize the local geometry. 
Rusu et al.~\cite{rusu2008aligning} use persistent Point Feature Histograms (PFH) to estimate a robust set of 16D features that describe the local pointwise geometry.
They~\cite{rusu2009fast} refine PFH and propose Fast Point Feature Histograms (FPFH), retaining most of PFH's discriminative power while improving computational efficiency.
However, the traditional registration methods typically require complicated optimization strategies, and may fail in complex scenes. 

Recently, many learning-based works~\cite{huang2021predator,yew2022regtr,ao2023buffer,mei2023overlap,liu2023density} about 3D descriptors have been proposed.
FCGF~\cite{choy2019fully} computes the features in a single pass through a fully convolutional neural network without relying on keypoint detection.
ImLoveNet~\cite{chen2022imlovenet} uses an intermediate misaligned 
image and multi-modal features to enhance low-overlap 
registration performance.
GeoTransformer~\cite{qin2023geotransformer} embeds the geometric features of point clouds into a Transformer to achieve robust superpoint matching.
PSReg~\cite{huang2025psreg} mitigates outlier matches caused by ambiguous features within overlapping regions.
However, supervised methods are constrained by their dependence on large paired datasets, limiting real-world applicability. 
In contrast, our framework enhances registration accuracy by utilizing learned geometric priors, offering a more scalable solution without extensive labeled data.

Compared to the supervised methods, unsupervised registration methods~\cite{wang2023zero,wimmer2024back, morreale2024neural} attract less attention.
BYOC~\cite{el2021bootstrap} learns visual and geometric features from RGB-D video without relying on ground-truth pose or correspondence.
UDPReg~\cite{mei2023unsupervised} leverages Gaussian Mixture Models to achieve self-supervised.
EYOC~\cite{liu2024extend} trains a feature extractor progressively, iterating from near to far to achieve self-supervision of distant point clouds.
Yuan et al.~\cite{yuan2024exploring} propose that self-supervised frameworks have unverified suitability for registration tasks.
Buffer-X~\cite{seo2025buffer} utilizes a zero-shot registration pipeline to improve robustness in diverse scenes.
However, these methods still require a substantial amount of data for self-supervised training.

\subsection{Outlier Removal Methods}
Although our method is fundamentally based on feature extraction, it extends beyond this by refining the registration results achieved by existing approaches, effectively serving as a refinement mechanism. 
This capability aligns closely with the objectives of outlier removal techniques, which aim to improve data integrity by eliminating inconsistencies. 
Therefore, we provide a brief introduction to outlier removal methods.
Traditional outlier filtering methods, such as RANSAC~\cite{fischler1981random} and its variants~\cite{barath2018graph, le2019sdrsac, li2020gesac}, use repeated sampling and verification for outlier rejection.
SC$^2$-PCR++~\cite{chen2023sc} proposes a second-order spatial compatibility metric to compute the similarity between correspondences. 
VBReg~\cite{jiang2023robust} develops a novel variational non-local network-based outlier rejection framework for robust alignment. 
MAC~\cite{zhang20233d} loosens the previous maximum clique constraint and mines more local consensus information in a graph for accurate pose hypotheses generation.
FastMAC~\cite{zhang2024fastmac} enhances MAC by introducing graph signal processing into the domain of correspondence graph, achieving real-time speed while leading to little performance drop.
Since higher-quality correspondences result in more accurate transformations, enhancing the power of the features becomes essential to achieve superior performance.

\begin{figure*}[ht]
  \centering
  \includegraphics[width=0.95\linewidth]{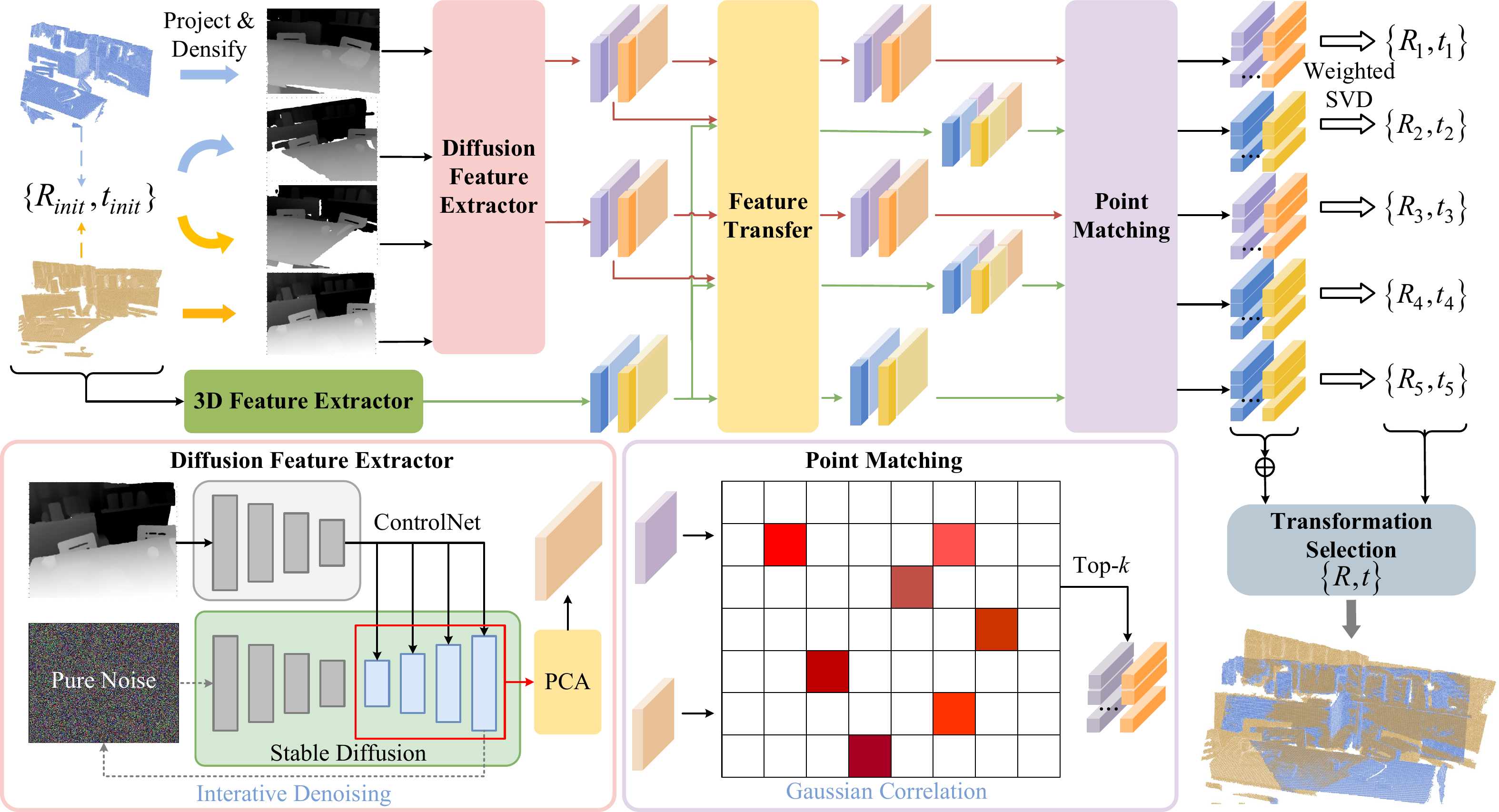}
  \caption{
  Pipeline of our framework. 
  Our framework begins by employing GeoTransformer~\cite{qin2023geotransformer} as a 3D feature extractor to derive an initial transformation for aligning the point clouds.
  The initial transformation is applied to project the point clouds into both the reference and source views, resulting in two pairs of depth images.
  Subsequently, a pre-trained diffusion model is utilized to extract depth diffusion features.
  Next, we use the Feature Transfer Module to establish common points between the depth diffusion features and the previously extracted 3D features, and the features of these common points are generated through a feature concatenation operation.
  To leverage both common and non-common features effectively, an inlier aggregation module is introduced. It integrates point matching, weighted SVD, and transformation selection to consolidate correspondences.
  Finally, this module identifies a set of high-confidence correspondences by analyzing common and non-common points correspondences, then computes the final rigid transformation.
}
\label{fig:pipline}
\end{figure*}

\subsection{Diffusion Feature}

Diffusion models have recently demonstrated remarkable performance in image generation tasks~\cite{Prafulla2021Diffusion, chitwan2022palette, rombach2022high}, capable of producing more diverse images and proven to be unaffected by mode collapse.
Building on this foundation, diffusion models have been extensively studied and applied to other vision tasks, such as object detection~\cite{chen2023diffusiondet} and image segmentation~\cite{chen2023generalist, jiang2018difnet, tan2022semantic}.
Recent studies~\cite{zhang2024tale, tang2023emergent} leverage diffusion features for representation learning and semantic matching across RGB images, effectively capturing object variations both within and across categories.
Additionally, FreeReg~\cite{wang2023freereg} explores the integration of diffusion features into image-to-point cloud registration tasks, demonstrating their effectiveness in learning cross-modal representations for image-to-point cloud alignment.
Morreale et al.~\cite{morreale2024neural} propose NSM to distill semantic matches from pre-trained vision models.
Wimmer et al.~\cite{wimmer2024back} first propose to back-project features from large pre-trained 2D vision models onto 3D shapes and employ them for this task.
ZeroMatch~\cite{Jiang2025CVPR} uses the powerful image representations of SD to enhance geometric descriptors for robust registration.
Inspired by these works, we leverage diffusion features in point cloud registration tasks to optimize existing registration methods, thereby enhancing their performance.
\section{Method} \label{sec:method}

\subsection{Overview}
Given a reference point cloud $P = \{p_i\in \mathbb{R}^3 | i=1,2,...,M\}$ and a source point cloud $Q = \{q_i\in \mathbb{R}^3 | i=1,2,...,N\}$, point cloud registration aims to compute a rigid transformation  $T = \{R, t\} $ that aligns the two partial point clouds, captured from different viewpoints of the same scene or object.
Here, $R \in SO(3)$ represents a 3D rotation, and $t \in \mathbb{R}^3$ represents a 3D translation.
Let $p^k$ and $q^k$ denote the key points in $P$ and $Q$, respectively, while $F_p$ and $F_q$ are the corresponding pointwise features generated by feature descriptors. 
An initial transformation $T_{init}$ is estimated using these pointwise 3D features.
Our proposed method refines the rigid transformation between $P$ and $Q$  by leveraging $F_p$, $F_q$, and $T_{init}$.
Our method is technically straightforward, as demonstrated in Figure \ref{fig:pipline}.

\subsection{Stable Diffusion and ControlNet}

\textbf{Preliminary.} 
Our depth map diffusion features are built on the ControlNet framework, which leverages diffusion models—parameterized Markov chains trained via variational inference.
These models operate through two Markov chain processes: a forward process and a reverse process.
In the forward process, Gaussian noise $(\epsilon)$ is incrementally added to the input image over multiple steps, gradually transforming the image into an unstructured noise representation.
At any time step $\ell$, the image during the forward diffusion process is defined as:
\begin{equation}
    x_\ell = \sqrt{\alpha_\ell}x_0 + \sqrt{1-\alpha_\ell}\epsilon
\end{equation}
Here, $\alpha_\ell$ is a coefficient controlling the noise schedule, as defined in~\cite{ho2020denoising}, and $\ell \in \text{range}(S)$.
In the reverse process, the diffusion model employs a UNet to iteratively denoise the unstructured noise image, ultimately reconstructing the output image. The UNet is trained using the noisy RGB image $x_\ell$ and the corresponding noise added during the forward process as input data. The overall objective of the reverse process can be expressed as:
\begin{equation}
    L_{DM} = \mathbb{E}_{x, \epsilon \sim \mathcal{N}(0,1),\ell}\left [ \left \| \epsilon -\epsilon_\theta (x_\ell, \ell)  \right \|_2^2  \right ] 
\end{equation}
This loss function $L_{DM}$ minimizes the difference between the predicted noise $\epsilon_\theta(x_\ell, \ell)$ and the actual noise $\epsilon$, enabling the UNet to effectively reconstruct images during the reverse diffusion process.
Building upon this foundation, Stable Diffusion incorporates an encoder $E$ and a decoder $D$, derived from VQGAN~\cite{esser2021taming}, to enable seamless conversion between pixel space and latent space.

Recent advancements like ControlNet extend this framework by introducing an additional encoder to process depth maps. The extracted depth features are then used to guide the reverse process of SD, enabling it to generate images that align coherently with the input depth map, starting from pure Gaussian noise. 
Inspired by these developments, we leverage ControlNet to extract diffusion features from depth maps, which are then used for matching purposes.

\textbf{Depth Diffusion Feature Extraction.} 
Given the reference and source point clouds, we first apply the initial transformation $T_{init}$, calculated using existing point cloud registration methods, to project the reference and source point clouds into depth maps from the perspective of the reference point cloud.
Conversely, we apply the inverse of the initial transformation to project the reference and source point clouds into depth maps from the perspective of the source point cloud. 
This process yields two pairs of depth maps: one from the reference's perspective and the other from the source's perspective.
For depth maps, as discussed in FreeReg, we apply traditional erosion and dilation operations to densify them before extracting diffusion features using ControlNet. These depth maps are then input into ControlNet, where an additional encoder encodes the depth maps into the latent space to guide the reverse process of Stable Diffusion. During Stable Diffusion reverse process, the pure Gaussian noise is progressively denoised through a specified number of sampling steps.
Finally, as mentioned in~\cite{zhang2024tale}, we aggregate feature maps from specific layers in the decoder of the U-Net to obtain deep diffusion features. 
{For feature extraction, we selected layers [0, 3, 6] based on previous work \cite{zhang2024tale, tang2023emergent, wang2023freereg} and verified their effectiveness through ablation experiments.}
However, simple concatenation of these feature maps often results in a high-dimensional fusion feature, which not only contains redundant information but also leads to excessive memory usage and computational overhead.
{To address this, we employ principal component analysis (PCA) to reduce the dimensionality of the feature maps for each layer.}
The reduced feature maps are then upsampled to the same resolution and combined to form the final depth diffusion features.

\subsection{Refine Using Deep Diffusion Features}
Although we have obtained diffusion features, it is challenging to directly integrate them with 3D features. 
Therefore, we design a series of module aggregating these two types of features to refine the initial transformation.

\textbf{Pixel-to-point Feature Transfer.} 
{The pixel-wise coordinates associated with diffusion-derived feature maps exhibit a geometric misalignment with 3D point-wise coordinate systems. 
This discrepancy stems from the inherent disparity of camera projection parameters and the stochastic sampling mechanisms of diffusion models.
To bridge this cross-modal representation mismatch and aggregate the depth diffusion features with the 3D features, we design a Feature Transfer module. 
}

{
For point cloud $P$, the Feature Transfer module takes three inputs: the 3D feature $F_p$, the reference view diffusion feature $F_p^{ref}$, and the source view diffusion feature $F_p^{src}$, with their  associated coordinates sets $coor_p$, $coor_p^{ref}$, and $coor_p^{src}$, respectively.
This module utilizes a nearest-neighbor matching to identify common points between $coor_p$ and $coor_p^{ref}$, which are points with Euclidean Distances below a predefined threshold and are generally considered the same point, denoted as $coor_p^{com1}$. 
Similarly, common points between $coor_p$ and $coor_p^{src}$ are identified and denoted as $coor_p^{com2}$. The same process is applied to point cloud $Q$ to obtain $coor_q^{com1}$ and $coor_q^{com2}$.
Then, to fuse the correspondences of different views robustly, we implement a two-stage feature aggregation strategy.
The core idea involves independently normalizing the two features to align their scales and distributions, followed by concatenating them.}
The fused feature is calculated as:
\begin{equation}
    FusedF_{\{p,q\}}^{\{ref,src\}} = Concat(\|F_{\{p,q\}}\|_2, \|F_{\{p,q\}}^{\{ref,src\}}\|_2)
\end{equation}
where $\|\cdot\|_2$ represents \textbf{L2} norm, 
{$\{\cdot,\cdot\}$ represents choosing between two options, } 
and $Concat(\cdot,\cdot)$ denotes concatenation along the feature dimension.

\textbf{Inlier Aggregation Module.}
While the Feature Transfer module enhances correspondence accuracy through aggregated 3D-diffusion features, two limitations persist:  
First, exclusive reliance on common points discards potentially valid correspondences;  
Second, in specific cases where the number of common points is limited, the results may display considerable variability. 
In extreme situations, such as when fewer than three common points are available, estimating the rigid transformation may become infeasible.
To address this, we introduce an Inlier Aggregation Module that reweights the confidence scores for each correspondence and obtains a robust registration result.

{
The module operates on five input feature sets: the fused common point features from the reference view $FusedF_p^{ref}$ and $FusedF_q^{ref}$, the fused common point features from the source view $FusedF_p^{src}$ and $FusedF_q^{src}$, the reference-view diffusion features $F_p^{ref}$ and $F_q^{ref}$, the source-view diffusion features  $F_p^{src}$ and $F_q^{src}$, and the 3D features $F_p$ and $F_q$.
Then, it builds five correspondence sets via feature similarity matching: $\mathcal{C}_{common}^{ref}$ (reference view common features), $\mathcal{C}_{common}^{src}$ (source view common features), $\mathcal{C}_{diffu}^{ref} $ (reference view diffusion features), $\mathcal{C}_{diffu}^{src}$ (source view diffusion features), $\mathcal{C}_{3d}$ (3D features).}
Finally, the transformations $T_i = \{R_i, t_i\}$ are computed for the five sets of correspondences.
It can be  determined in closed form using weighted singular value decomposition (SVD):
\begin{equation}
    R_i, t_i =  \min_{R, t} {\textstyle \sum_{(p_{x_j}, q_{y_j})\in \mathcal{C}_i }w_j^i\left \| R\cdot p_{x_j}+t - q_{y_j}  \right \|_2^2 } 
\end{equation}
where $\mathcal{C}_i$ belong to five correspondence sets,  $w_j^i$ is confidence score for each correspondence.
Then, we select the transformation that allows the most inlier matches:
\begin{equation}
    R, t = \max_{R_i, t_i} {\textstyle \sum_{(p_{x_j}, q_{y_j})\in \mathcal{C}} \llbracket \left \| R_i\cdot p_{x_j}+t_i - q_{y_j}  \right \|_2^2 < \tau_a \rrbracket} 
\end{equation}
where $\llbracket \cdot \rrbracket$ is the Iverson bracket and $\tau_a$ is the acceptance radius. 
We then iteratively re-weight the confidence scores based on the surviving inlier matches, refining the transformation $R$, $t$ for $Nr$ iterations using weighted SVD.
Finally, the $K$ highest-scoring correspondences are used to estimate the final transformation. 

\section{Experiments}

\subsection{Experimental Setup}

\textbf{Datasets.}
We consider three datasets, i.e., the scene-scale indoor datasets 3DMatch \& 3DLoMatch~\cite{zeng20173dmatch}, and the scene-scale outdoor dataset KITTI~\cite{andreas2012are}.
3DMatch is a collection of 62 scenes, from which we use 8 for testing. 
Official 3DMatch dataset considers only scan pairs with $>30\%$ overlap. 
For the 3DLoMatch, It considers only scan pairs with overlaps between 10 and 30\% from 3Dmatch.
KITTI contains 11 sequences of LiDAR-scanned outdoor driving scenarios. 
We follow~\cite{choy2019fully} and use 8-10 for testing.

\textbf{Evaluation Criteria.}
Following~\cite{huang2021predator, yew2020rpm}, we evaluate the performance with three metrics: 
(1) Registration Recall (RR), i.e., the fraction of point cloud pairs whose transformation error is smaller than a certain threshold;
(2) Relative Rotation Error (RE), i.e., geodesic distance between estimated and GT rotation matrices;
(3) Relative Translation Error (TE), i.e., Euclidean distance between the estimated and GT translations.

\begin{figure*}[htbp]
  \centering
  \includegraphics[width=0.95\linewidth]{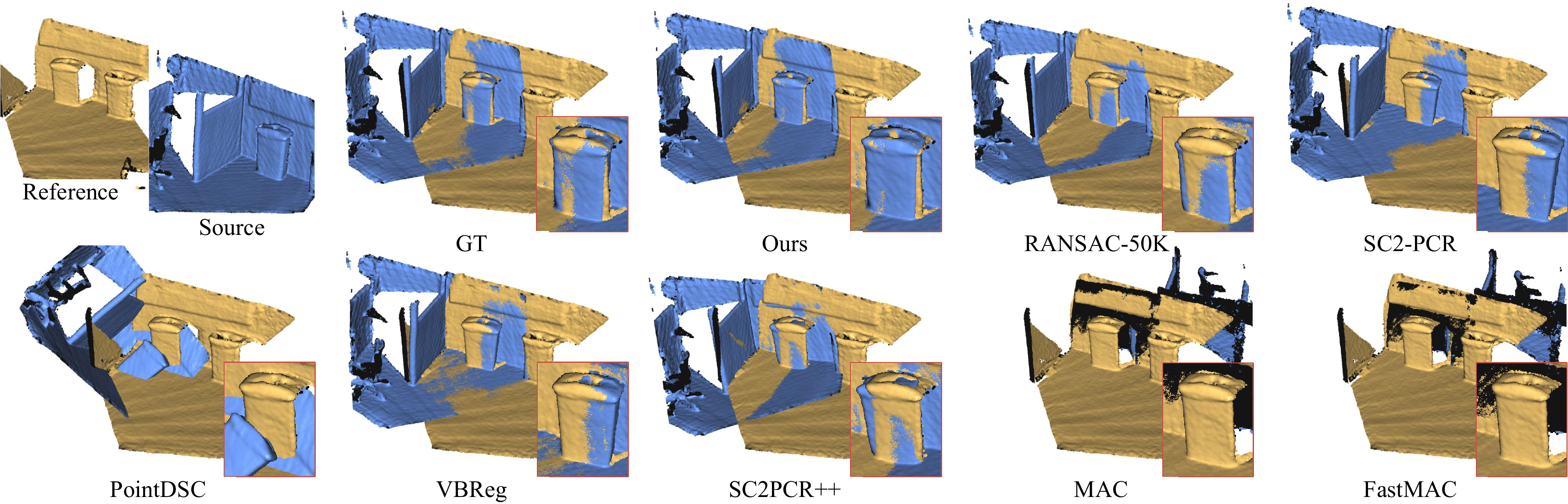}
  \caption{Registration visualization results on 3DMatch.}
  \label{fig:3dmatch}
\end{figure*}

\textbf{Implementation Details.}
While our approach and traditional outlier removal methods commonly employed in point cloud registration share an optimization-based framework, our method is fundamentally distinct in its underlying principles. 
Nonetheless, the input-output characteristics of the two approaches exhibit significant similarities.
Therefore, we compare our method with the baseline outlier removal methods. 
{To ensure a fair comparison, we utilize Geotransformer~\cite{qin2023geotransformer} as a initial method---a widely applied method in point cloud registration---to generate correspondences, 3D features, and initial transformations. }
Regarding the number of sampling points, we have struck a balance between efficiency and effectiveness by sampling 500 points for optimization on the 3DMatch dataset and setting 1000 points for the KITTI dataset. 
Since the input requirements of our method differ from those of outlier removal methods, for a fair comparison, we compute the initial transformation on the same sampling points. 
This typically implies that the initial transformation results may be inferior to those obtained by the prior methods themselves.

\textbf{Comparison methods.} 
{We evaluate both traditional and deep-learning-based approaches, including
RANSAC~\cite{fischler1981random}, 
SC$^2$-PCR~\cite{chen2022sc2},
PointDSC~\cite{bai2021pointdsc},
SC$^2$-PCR++~\cite{chen2023sc}, VBReg~\cite{jiang2023robust},
MAC~\cite{zhang20233d} and FastMAC~\cite{zhang2024fastmac}.}

\begin{table}[htbp]
  \centering
  \caption{Registration results on 3DMatch dataset.}
  \label{tab:3dmatch}
 \begin{tabular}{l|ccc}
\hline
Method($\dagger$:Learning-based)          & RR$\uparrow $       & RE$\downarrow $      & TE$\downarrow $            \\ \hline
RANSAC-50K~\cite{fischler1981random}    &  91.4 &   2.24  &   0.071    \\ 
RANSAC-1M~\cite{fischler1981random}    & 91.6  & 2.24    &  0.070     \\ 
SC$^2$-PCR~\cite{chen2022sc2}    &   93.1   &   1.86  &   0.052         \\ 
PointDSC$^\dagger$~\cite{bai2021pointdsc}        &  92.3   &   1.75   &   0.051       \\ 
VBReg$^\dagger$~\cite{jiang2023robust}            &  94.1   &  1.87  &    0.052      \\  
SC$^2$-PCR++~\cite{chen2023sc}    &   91.2  &   1.84    &    0.052         \\ 
MAC~\cite{zhang20233d}            & 93.8   &  1.60  &    0.052                 \\ 
FastMAC~\cite{zhang2024fastmac}          &  93.8  & 1.86   &   0.065           \\ \hline
Ours              &  \textbf{94.5}  &  \textbf{1.56}     &   \textbf{0.050}      \\\hline
\end{tabular}
\end{table}

\begin{table}[htbp]
  \centering
  \caption{Registration results on 3DLoMatch dataset.}
  \label{tab:3dlomatch}
 \begin{tabular}{l|ccc}
\hline
 Method ($\dagger$:Learning-based)                  & RR$\uparrow $       & RE$\downarrow $      & TE$\downarrow $            \\ \hline
RANSAC-50K~\cite{fischler1981random} &  65.9    &    3.62     &    0.101        \\ 
RANSAC-1M~\cite{fischler1981random} &   66.4   &  3.71   &      0.105           \\ 
SC$^2$-PCR~\cite{chen2022sc2}    &  76.7    &   2.74    &    0.077           \\ 
PointDSC$^\dagger$~\cite{bai2021pointdsc}       &    69.8   &    2.60     &     0.075           \\ 
VBReg$^\dagger$~\cite{jiang2023robust}            &  \textbf{77.7}   &    2.68     &    0.075     \\ 
SC$^2$-PCR++~\cite{chen2023sc}     &    71.8      &    2.69     &   0.077              \\ 
MAC~\cite{zhang20233d}         &  76.4  & 2.57 &   0.075         \\ 
FastMAC~\cite{zhang2024fastmac}  & 75.4   & 2.92   &  0.085        \\\hline
Ours              & 76.7  &  \textbf{2.55}    &   \textbf{0.074}     \\\hline
\end{tabular}
\end{table}

\subsection{Indoor Benchmarks: 3DMatch \& 3DLoMatch}
\begin{figure*}[htbp]
  \centering
  \includegraphics[width=0.95\linewidth]{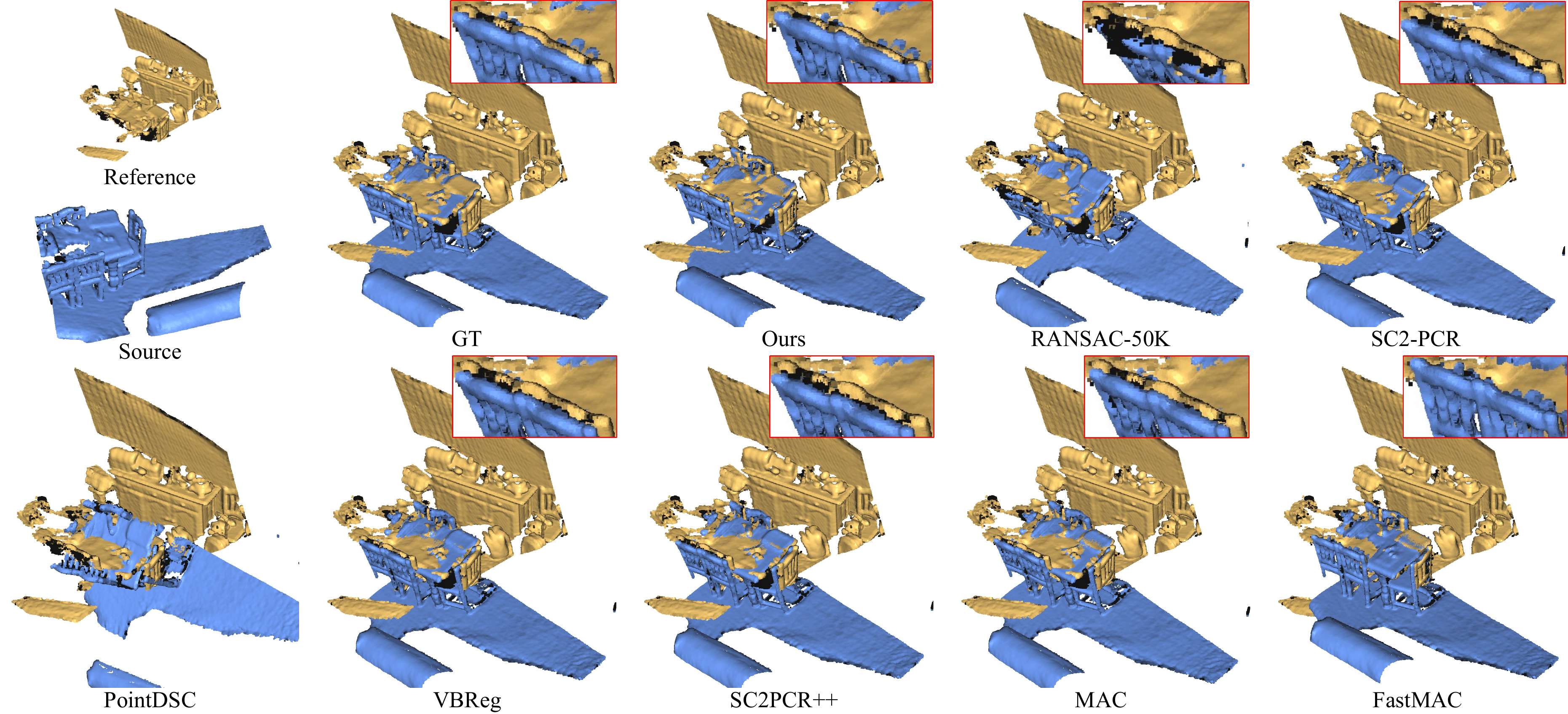}
  \caption{Registration visualization results on 3DLoMatch.}
  \label{fig:3dlomatch}
\end{figure*}

The visual results presented in Figures ~\ref{fig:3dmatch} and ~\ref{fig:3dlomatch}, combined with the quantitative results summarized in Tables ~\ref{tab:3dmatch} and \ref{tab:3dlomatch}, support the following conclusions:
(1) Both the geometric-only and deep-learned methods are outperformed by our method on the 3DMatch and 3DLoMatch datasets, demonstrating its ability to effectively optimize point cloud registration for indoor scenes. This underscores its proficiency in enhancing the registration capabilities of point cloud methods.
(2) Even when compared to deep learning approaches, our method achieves superior performance without any data training, thereby proving its robust generalization capabilities.
(3) In addition to the RR metric, we also attain the best results in terms of RE and TE metrics.
{
It is worth noting that in Table~\ref{tab:3dmatch} and ~\ref{tab:3dlomatch}, we use Geotransformer to generate the initial transformation and 3D features. 
For other refinement methods in the tables, we also use Geotransformer to generate their initial correspondences.
}

\textbf{Boosting other supervised methods.}  To further evaluate the effectiveness and generalization ability of our method, we selected several classic learning-based point cloud registration methods as baseline approaches for comparison. The considered methods include Predator~\cite{huang2021predator}, CoFiNet~\cite{yu2021cofinet}, and GeoTransformer~\cite{qin2023geotransformer}. 

As shown in Table~\ref{tab:boost}, we have achieved significant improvements in registration recall rate, rotation error, and translational error across all tested methods on the 3DMatch dataset, demonstrating robustness and generalization capabilities. 
Notably,  both Predator and CoFiNet outperformed GeoTransformer after being optimized by our method. 
Building upon GeoTransformer, we attain a registration recall rate of 94.5\% on the 3DMatch dataset.
Furthermore, we achieve further enhancements in the rotation error and translational error metrics, demonstrating the ability to refine samples with accurate registrations.

\begin{table}[htbp]
  \centering
  \caption{Performance boosting for deep-learned methods when combined with our method.}
  \label{tab:boost}
 \begin{tabular}{c|ccc}
\hline
\multirow{2}{*}{} & \multicolumn{3}{c}{3DMatch}  \\
                  & RR$\uparrow $       & RE$\downarrow $      & TE$\downarrow $            \\ \hline
FCGF~\cite{choy2019fully}       &    67.9    &   2.20     &   0.078         \\ 
Predator~\cite{huang2021predator}     &  89.0    &  2.02    & 0.064           \\  
CoFiNet~\cite{yu2021cofinet}       &  89.3   & 2.44 &  0.067 \\  
GeoTransformer~\cite{qin2023geotransformer} &  92.0 &   1.72   & 0.062  \\  
\hline
\multirow{2}{*}{FCGF+Ours} & 76.2   &  2.07   &  0.055   \\
                 &   8.3 $\uparrow$  &  0.13 $\downarrow$  &  0.023 $\downarrow$         \\ \cdashline{1-4}   
\multirow{2}{*}{Predator+Ours} &   92.7   & 1.63    &  0.051    \\
   & 4.7 $\uparrow$  &  0.39 $\downarrow$  &  0.013 $\downarrow$ \\  \cdashline{1-4}
\multirow{2}{*}{CoFiNet+Ours} &   93.0  &  1.60      &   0.051   \\
     & 3.7 $\uparrow$  &  0.84 $\downarrow$  &  0.016 $\downarrow$ \\ \cdashline{1-4} 
\multirow{2}{*}{GeoTransformer+Ours}    &  94.5  &  1.56     &   0.050      \\
        & 2.5 $\uparrow$  &  0.16 $\downarrow$  &   0.012 $\downarrow$ \\\hline
\end{tabular}
\end{table}

\begin{table}[htbp]
  \centering
  \caption{Registration results on KITTI dataset.}
  \label{tab:kitti}
 \begin{tabular}{c|ccc}
\hline
Method ($\dagger$:Learning-based)                   & RR$\uparrow $       & RE $\downarrow $     & TE  $\downarrow $          \\ \hline
Geotransformer~\cite{qin2023geotransformer}   &     \textbf{99.8}  &   \textbf{0.24}       &    0.068     \\
RANSAC-50K~\cite{fischler1981random}           &    \textbf{99.8}    &    0.42     &       0.069          \\
SC$^2$-PCR~\cite{chen2022sc2}  &    \textbf{99.8}      &   0.57      &      0.105            \\ 
PointDSC$^\dagger$~\cite{bai2021pointdsc}       &   99.3   &  0.87   &      0.176          \\ 
VBReg$^\dagger$~\cite{jiang2023robust}           &   \textbf{99.8}   &  0.57       &    0.115              \\ 
MAC~\cite{zhang20233d}   &  99.6  &   0.39   &    0.128   \\ 
FastMAC~\cite{zhang2024fastmac}    &     99.5     &    0.38     &    0.120             \\ \hline
Ours             & \textbf{99.8 }  &   \textbf{0.24  }  &  \textbf{ 0.064}      \\\hline
\end{tabular}
\end{table}

\subsection{Outdoor Benchmark: KITTI odometry}
We have also tested our method with other SOTA methods on the KITTI odometry dataset. As shown in Table~\ref{tab:kitti}, for the three metrics of RR, RE, and TE, the performance of certain outlier removal methods has declined to some extent due to GeoTransformer already approaching their performance limits. 
Although our method is also unable to further enhance the registration recall rate, it can still improve the rotation error and translational error metrics without compromising the effectiveness of the original methods.
{Consistent with Tables ~\ref{tab:3dmatch} and ~\ref{tab:3dlomatch}, Table ~\ref{tab:kitti} also uses Geotransformer as the initialization method.}

\subsection{Ablation Experiments}
We conduct ablation studies and analytical experiments on the 3DMatch dataset. We independently assess the contribution of each methodological component introduced in Section 3, with results detailed in Tables ~\ref{tab:layer} and ~\ref{tab: ablation}.
More ablation studies are provided in the supplementary materials.

\textbf{Diffusion layer selection.}
In table ~\ref{tab:layer}, we report the size of output feature maps of 13 layers in the UNet of Stable Diffusion. The feature map size is divided into four levels, i.e. small group (8 $\times$ 11, layers 0-2), medium group (16 $\times$ 22, layers 3-5), large group (32 $\times$ 44, layers 6-8), and extra-large group (64 $\times$ 88, layers 9-12). To better assess the contribution of each layer to the depth map matching process, we conduct an experiment where Principal Component Analysis (PCA) was excluded, relying solely on diffusion features to estimate correspondences. 
The transformation errors obtained from the derived correspondences are summarized in the table. 
The experimental results demonstrate that integrating features from layers 0, 3, and 6 leads to a significant improvement in registration performance. 
Conversely, incorporating features from layer 9 adversely affects the registration process, indicating a potential limitation of extra-large layer features in this context.

\begin{table}[htbp]
  \centering
  \caption{Layer selection in diffusion feature extraction.}
  \label{tab:layer}
\begin{tabular}{ccccc}
\hline
\multirow{2}{*}{Layer} & \multirow{2}{*}{Feature Map} & \multicolumn{3}{c}{3DMatch} \\
     &      & RR  $\uparrow $    & RE $\downarrow $     & TE  $\downarrow $    \\ \hline
0           & $8\times11$  &     54.2    &     19.02    &    1.772     \\
3        &  $16\times22$   &    70.7     &    16.45     &    0.554     \\

6          &                   $32\times44$                                  &        67.7 &     17.00    &    0.505     \\
9       &      $64\times88$          &    28.2     &     40.84    &  1.128
\\ \hline
{[}0,3{]}  &    $16\times22$    & 69.8      &    15.60     &   0.582         \\
{[}0,6{]}   &      $32\times44$      &    74.6     &    16.03    &  0.476       \\
{[}0,9{]}                                           &   $64\times88$                                                  &      53.4   &   26.36      &    0.735     \\
\hline
{[}0,3,6{]}     &    $32\times44$         & \textbf{76.0}  &   \textbf{14.67}  &  \textbf{0.442}   \\
{[}0,3,9{]}        &    $64\times88$       &    60.9     &    21.90     &  0.624       \\
\hline
{[}0,3,6,9{]}                                         &     $64\times88$                                                & 63.3  &   20.33  &   0.600        \\ \hline
\end{tabular}
\end{table}

\textbf{The effectiveness of components.} We conduct an ablation study to evaluate the contribution of each component in our framework, as summarized in Table~\ref{tab: ablation}. 
The PCA module exhibited minimal impact on overall performance, which is consistent with its primary purpose of reducing the dimensionality of diffusion features to alleviate computational overhead. 
In contrast, the Inlier Aggregation module demonstrated the most significant contribution to the method’s effectiveness. 
While diffusion features are successful in generating a large number of correct correspondences, they are prone to introducing outliers. 
The Inlier Aggregation module addresses this issue by refining these correspondences, ensuring greater accuracy. 
Additionally, the feature transfer and feature fusion operations were found to be crucial for the overall performance, underscoring their importance in enhancing the method's robustness and efficacy.

\begin{table}[htbp]
  \centering
  \caption{Ablation of Components.}
  \label{tab: ablation}
 \begin{tabular}{c|ccc}
\hline
\multirow{2}{*}{} & \multicolumn{3}{c}{3DMatch}  \\
                  & RR $\uparrow $     & RE $\downarrow $     & TE  $\downarrow $          \\ \hline

Geotransformer~\cite{qin2023geotransformer}   &   92.0  &   1.72     &    0.062    \\
w/o PCA    &  94.1   &    1.56     &  0.052    \\
w/o Feature Transfer   &   93.1     &    1.65     &   0.051    \\

w/o Inlier Aggregation   & 92.8  &  1.69  &    0.060     \\ 
Ours              & \textbf{94.5 }  &   \textbf{1.56}  &  \textbf{ 0.050}      \\\hline
\end{tabular}
\end{table}
\section{Conclusion}
In this paper, we introduce a novel zero-shot approach to enhance point cloud registration by incorporating diffusion-based features extracted from depth images. Inspired by recent advances in leveraging pre-trained diffusion models for semantic correspondence in images, our method refines traditional point cloud registration techniques by combining depth diffusion features with geometric features. 
This fusion improves the accuracy of correspondences between misaligned point clouds without requiring a training dataset. Extensive experiments on multiple datasets demonstrate that our approach significantly enhances the performance of existing registration methods and exhibits robust generalization capabilities across various scenarios. This work opens up new possibilities for more effective and efficient point cloud registration, particularly in real-world applications where labeled data may be scarce.

\section*{Acknowledgement} 
This work was supported by the National Natural Science Foundation of China (No. T2322012, No. 62172218), the Shenzhen Science and Technology Program (No. JCYJ20220818103401003, No.
JCYJ20220530172403007), and the Shenzhen Longhua Science and Technology Innovation Special Funding Project (Industrial Sci-Tech Innovation Center of Low-Altitude Intelligent Networking).

{
    \small
    \bibliographystyle{ieeenat_fullname}
    \bibliography{main}
}

\end{document}